\documentclass[10pt,twocolumn,letterpaper]{article}

\usepackage{wacv}
\usepackage{times}
\usepackage{epsfig}
\usepackage{graphicx}
\usepackage{amsmath}
\usepackage{amssymb}
\usepackage{subcaption}
\usepackage{multirow}
\usepackage{numprint}
\def\wacvPaperID{699} 

\wacvfinalcopy 

\ifwacvfinal
\def\assignedStartPage{1} 
\fi

\ifwacvfinal
\usepackage[breaklinks=true,bookmarks=false]{hyperref}
\else
\usepackage[pagebackref=true,breaklinks=true,colorlinks,bookmarks=false]{hyperref}
\fi

\ifwacvfinal
\else
\fi

\begin{document}

\title{Seeing Implicit Neural Representations as Fourier Series}

\author{Nuri Benbarka\thanks{nuri.benbarka@uni-tuebingen.de (equal contribution)}, Timon H\"ofer\thanks{timon.hoefer@uni-tuebingen.de (equal contribution)}, Hamd ul-moqeet Riaz, Andreas Zell\\
	University of T\"ubingen\\
	Wilhelm-Schickard-Institute for Computer Science, Sand 1, 72076 T\"ubingen\\
	{\tt\small firstname.lastname@uni-tuebingen.de}
}

\maketitle

\begin{abstract}
	Implicit Neural Representations (INR) use multilayer perceptrons to represent high-frequency functions in low-dimensional problem domains. Recently these representations achieved state-of-the-art results on tasks related to complex 3D objects and scenes. A core problem is the representation of highly detailed signals, which is tackled using networks with periodic activation functions (SIRENs) or applying Fourier mappings to the input. This work analyzes the connection between the two methods and shows that a Fourier mapped perceptron is structurally like one hidden layer SIREN. Furthermore, we identify the relationship between the previously proposed Fourier mapping and the general d-dimensional Fourier series, leading to an integer lattice mapping. Moreover, we modify a progressive training strategy to work on arbitrary Fourier mappings and show that it improves the generalization of the interpolation task. Lastly, we compare the different mappings on the image regression and novel view synthesis tasks. We confirm the previous finding that the main contributor to the mapping performance is the size of the embedding and standard deviation of its elements.
\end{abstract}

\section{Introduction}
INR is a novel field of research in which the traditional discrete signal representation (e.g., images as discrete grids of pixels, 3D shapes as voxel grids or meshes) are replaced with continuous functions that map the input domain of the signal (e.g., coordinates of a specific pixel in the image) to a representation of color, occupancy or density at the input location. However, these functions typically are not analytically tractable; INRs approximate those functions with fully connected neural networks (also called multi-layer perceptrons (MLPs)).
\begin{figure}
	\begin{center}
		\includegraphics[trim={6cm 1.5cm 26cm 1.5cm},clip,width=0.5\textwidth]{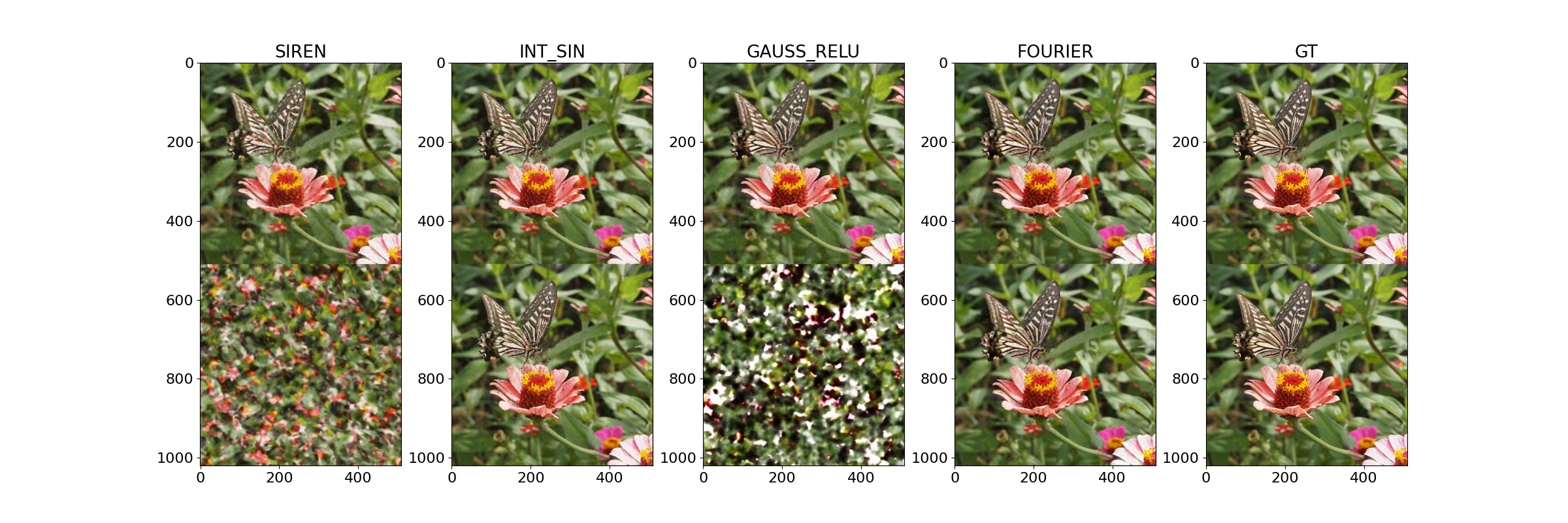}
	\end{center}
	\caption{Visualization of SIREN \cite{sitzmann2020implicit}, Gauss ReLU \cite{tancik2020fourier} and our method (INT\_SIN). The top row is the first period and bottom row is the second period, which shows our method enforcing periodicity.}
	\label{fig:main}
\end{figure}

INRs are not coupled to the spatial resolution (e.g., voxel size in a 3D scene) and theoretically have infinite resolution. Therefore, these representations are naturally suited in applications with high-dimensional signals and heavy memory consumption. Also, since they are differentiable, they are suitable for gradient-based optimization and machine learning. In addition, the application of INRs for images \cite{henzler2020learning,stanley2007compositional}, volume density \cite{mildenhall2020nerf}, and occupancy \cite{mescheder2019occupancy} enhanced the performance on various tasks such as shape representation \cite{chen2019learning,deng2019nasa,genova2019learning,genova2020local,jiang2020local,michalkiewicz2019implicit,park2019deepsdf}, texture synthesis \cite{henzler2020learning,oechsle2019texture}, and shape inference from images \cite{liu2020dist,liu2019learning}.

However, early architectures lacked accuracy in high-frequency details. Sitzmann et al. \cite{sitzmann2020implicit} proposed SIRENs, which could represent high frequencies. They argued that sinusoidal activations work better than ReLU networks because ReLU networks are piecewise linear, and their second derivative is zero. As a result, they are incapable of modeling data contained in higher-order derivatives of signals. However, a concurrent work \cite{mildenhall2020nerf} proposed positional encoding, which also enabled the networks to learn high-frequency information. The positional encoding uses a heuristic sinusoidal mapping to input coordinates before passing them through a ReLU network. They did follow-up work \cite{tancik2020fourier} exploring the general Fourier mapping and explaining why it worked using a Neural Tangent Kernel (NTK) framework \cite{jacot2018neural}. They found out that the Fourier mapping transforms the NTK into a shift-invariant kernel. And modifying the mapping parameters enables tuning the NTK's spectrum, therefore controlling the range of frequencies the network can learn. They also showed that a random Fourier mapping with low standard deviation learns only low frequencies of the signal. On the contrary, a high standard deviation lets the network learn high frequencies only, which leads to over-fitting. They recommended a linear search to find the optimal value of the standard deviation for the corresponding task. They also showed that increasing the number of parameters in the mapping improves the performance constantly.

However, we wonder what the difference between SIRENs and Fourier mapping is? 
Will the performance be saturated when we continue to increase the mapping parameters?   
Is there a way to avoid over-fitting when training networks using Fourier mapping?
And is random Fourier mapping the optimal mapping? 

To answer these questions, we explored the mathematical connection between Fourier mappings and SIRENs and showed that a Fourier mapped perceptron is structurally like a one hidden layer SIREN. However, in the SIREN case, the mapping is trainable, and it is represented in the amplitude-phase form instead of the sine-cosine form in the case of Fourier mappings.

Also, we looked at the functions we want to learn, and we observed that they have a limited input domain (e.g., the height and width of an image), and their values are defined on a finite set. Hence, we can assume that they are continuous and periodic over their input bound, which satisfies all the requirements to represent them with a Fourier series. Furthermore, we determined the \textit{d}-dimensional Fourier series's trigonometric form and showed that it is precisely a single perceptron with an integer lattice mapping applied to its inputs. The weights of that perceptron are the Fourier series coefficients. As the Fourier series can theoretically represent any periodic signal, this perceptron can represent any periodic signal if it has an infinite number of frequencies in its mapping. However, in practice, the Fourier series coefficients are finite, and we can get them by sampling the signal at the Nyquist rate (twice the bandwidth) and applying a fast Fourier transform (FFT) to the signal. Thus, the number of Fourier coefficients is the theoretical upper bound of the number of parameters needed in the mapping.

Moreover, we modified the progressive training strategy of \cite{lin2021barf}, where we train the lower frequencies in the initial training phase and gradually add the higher frequency components as the training progresses. As a result, we show that our \textit{Progressive Training} strategy avoids the problem of over-fitting. Finally, we tested our proposed \textit{Integer Lattice} mapping in the image regression and novel view synthesis tasks. We found out that the main contributor to the mapping performance is the number of parameters and the standard deviation, as was shown in \cite{tancik2020fourier}. In summary, we offer the following contributions:
\begin{description}
	\item[$ \bullet$ ] We introduce an integer Fourier mapping and prove that a perceptron with this mapping is equivalent to a Fourier series.
	\item[$ \bullet$ ] We explore the mathematical connection between Fourier mappings and SIRENs and show that a Fourier mapped perceptron is structurally like a one hidden layer SIREN.
	\item[$ \bullet$ ] We show that the integer mapping forces periodicity of the network output.
	\item[$ \bullet$ ] We modify the progressive training strategy of \cite{lin2021barf} and show that it improves the generalization of the interpolation task.
	\item[$ \bullet$ ] We compare the different mappings on the image regression and novel view synthesis tasks and verify the previous findings of \cite{tancik2020fourier} that the main contributor to the mapping performance is the number of elements and standard deviation.
\end{description}

\section{Related work}
Inspired by INRs' recent success, by outperforming grid-, point- and mesh-based representations (for the first time in 2018 \cite{park2019deepsdf},\cite{mescheder2019occupancy},\cite{chen2019net}), many works based on INRs achieved state-of-the-art results in 3D computer vision \cite{atzmon2020sal,gropp2020implicit,jiang2020local,peng2020convolutional,chabra2020deep,sitzmann2020implicit}. Moreover, impressive results are obtained across different input domains, e.g., from 2D supervision \cite{sitzmann2019scene,niemeyer2020differentiable,mildenhall2020nerf}, 3D supervision \cite{saito2019pifu,Oechsle2019ICCV}, to dynamic scenes \cite{Niemeyer2019ICCV} which can be represented by space-time INR. 

In early architectures, there was a lack of accuracy in fine details of signals. Mildenhall et al. \cite{mildenhall2020nerf} proposed positional encodings to tackle this problem, then Tancik et al. \cite{tancik2020fourier} further explored positional encodings in an NTK framework, showing that mapping input coordinates to a representation close to the actual Fourier representation before passing them to the MLP lead to a good representation of the high-frequency details. Furthermore, they showed that random Fourier mappings achieved superior results than if one takes the simple positional encoding. Sitzmann et al. \cite{sitzmann2020implicit} also attempted to solve the problem of getting high-frequency details. They proposed SIRENs and demonstrated that SIRENs are suited for representing complex signals and their derivatives. In both solutions, they used a variant of Fourier neural networks (FNN) for the first layer of the MLP. FNN are neural networks that use either sine or cosine activations to get their features \cite{liu2013fourier}. 

The first attempt to build an FNN was by \cite{gallant1988there}. They proposed a one-layer hidden neural network with a cosine squasher activation function and showed if they hand-wire certain weights, it will represent a Fourier series. Silvescu  \cite{silvescu1999fourier} proposed a network that did not resemble a standard feedforward neural network. However, they used a cosine activation function to get the features. Liu et al.   \cite{liu2013fourier} introduced the general form for Fourier neural networks in a feedforward manner. They also proposed a strategy to initialize the frequencies of the embedding, which helped for convergence. Our work will show another way to initialize the embedding, which results in a neural network that is precisely a Fourier series.


\section{Method}

\subsection{Integer lattice mapping}
\label{sec:integer lattice mapping}
This section explains how a perceptron with an integer lattice Fourier mapping applied to its inputs is equivalent to a Fourier series. First, we present the Fourier mapped perceptron equation and then link it to the Fourier series's general equation. The fundamental building block of any neural network is the perceptron, and it is defined as
\begin{align}
	\label{eq:perceptron}
	\textbf{y}(\textbf{x}, \textbf{W}', \textbf{b}) = g(\textbf{W}'\cdot \textbf{x}  + \textbf{b}).
\end{align}
Here $ \textbf{y} \in \mathbb{R}^{d_{out}} $ is the perceptron's output, $ g(\cdot) $ is the activation function (usually non-linear), $ \textbf{x} \in \mathbb{R}^{d_{in}} $ is the input, $ \textbf{W}' \in \mathbb{R}^{d_{out}\times d_{in}} $ is the weight matrix, and $ \textbf{b} \in \mathbb{R}^{d_{out}} $ is the bias vector. Now, if we let $ g(\cdot) $ to be the identity function and apply a Fourier mapping to the input we get
\begin{align}
	\label{eq:FourierPerceptron}
	\textbf{y}(\textbf{x}, \textbf{W}) = \textbf{W}\cdot \gamma(\textbf{x})  + \textbf{b},
\end{align}
where $ \gamma(x) $ is the Fourier mapping defined as
\begin{align}\label{eq:fouriertransform}
	\begin{split}
		\gamma(\textbf{x}) = \left(\begin{array}{ccc}
			\cos(2\pi \textbf{B}\cdot \textbf{x}) \\
			\sin(2\pi \textbf{B}\cdot \textbf{x}) \\
		\end{array}\right).
	\end{split}
\end{align}

$\textbf{W}\in \mathbb{R}^{d_{out} \times 2m}$, $ \textbf{B} \in \mathbb{R}^{m \times d_{in}}$ is the Fourier mapping matrix, and $ m $ is the number of frequencies. Equation \ref{eq:FourierPerceptron} is the general equation of a Fourier mapped perceptron, and we will relate it to the Fourier series's general equation.

A Fourier series is a weighted sum of sines and cosines with incrementally increasing frequencies that can reconstruct any periodic function when its number of terms goes to infinity. In applications that use coordinate-based MLPs, the functions we want to learn are not periodic. However, their inputs are naturally bounded (e.g., height and width of an image). Accordingly, it doesn't harm if we assume that the input is periodic over its input's bounds to represent it as a Fourier series. We will explain later why this assumption has many advantages. A function $ f:\mathbb{R}^{d_{in}}\to \mathbb{R}^{d_{out}}$ is periodic with a period $ p\in\mathbb{R}^{d_{in}} $ if
\begin{align}
	\label{eq:periodic}
	f(\textbf{x}+\textbf{n}\circ \textbf{p})= f(\textbf{x}) \;\;\;\; \forall \textbf{n} \in \mathbb{Z}^d,
\end{align}
where $\circ$ is the Hadamard product. As it is plausible to normalize the inputs to their bounds, we assume that each variable's period is 1. The Fourier series expansion of function \eqref{eq:periodic} with $ \textbf{p} = \mathbf{1}^d $ is defined by \cite{osgood2019lectures}:

\begin{align}
	\label{eq:complex}
	f(\textbf{x}) =  \displaystyle\sum_{\textbf{n}\in\mathbb{Z}^d} c_\textbf{n}e^{2\pi i \textbf{n}\cdot\textbf{x}},
\end{align}

where $ c_\textbf{n} $ are the Fourier series coefficients, and they are calculated by:
\begin{align}
	c_\textbf{n} = \int_{[0,1]^d} f(\textbf{x}) e^{-2\pi i \textbf{n} \textbf{x}}  d\textbf{x}.
\end{align}

For real-valued functions, it holds that $ c_\textbf{n} = c_\textbf{-n}^* $ where $ c_\textbf{n}^* $ is the conjugate of $ c_\textbf{n} $. Using Euler's formula and mathematical induction we showed that equation \eqref{eq:complex} can be written as (see supplementary material):
\begin{align}
	f(\textbf{x}) =  \displaystyle\sum_{\textbf{n}\in\mathbb{N}_0\times\mathbb{Z}^{d-1}} a_\textbf{n}\cos({2\pi  \textbf{n}\cdot\textbf{x}}) + b_\textbf{n}\sin({2\pi  \textbf{n}\cdot\textbf{x}})\label{eq:Fourierseries}
\end{align}
\begin{align}
	\begin{split}
		a_\textbf{0}&=c_\textbf{0},\\  a_\textbf{n}&=\begin{cases}
			0 \; \;  \exists j \in \{2,\dots,d\}:\, n_1=\dots=n_{j-1}=0 \wedge n_j<0\\2 \text{Re}(c_\textbf{n})\; \; \; \text{otherwise},\end{cases}\\
		b_\textbf{n}&=\begin{cases}
			0 \; \;  \exists j \in \{2,\dots,d\}:\, n_1=\dots=n_{j-1}=0 \wedge n_j<0\\-2 \text{Im}(c_\textbf{n})\; \; \; \text{otherwise}.
		\end{cases}
	\end{split}
\end{align}
And if we write equation \eqref{eq:Fourierseries} in vector form, we get
\begin{align}
	\label{eq:FourierSeriesVector}
	f(\textbf{x}) =  (a_\textbf{B},b_\textbf{B})\cdot\left(\begin{array}{ccc}
		cos(2\pi \textbf{B}\cdot \textbf{x}) \\
		sin(2\pi \textbf{B}\cdot \textbf{x}) \\
	\end{array}\right),
\end{align}
where $a_\textbf{B}=(a_\textbf{n})_{\textbf{n}\in \textbf{B}}$, and $b_\textbf{B}=(b_\textbf{n})_{\textbf{n}\in \textbf{B}}$.
Now, if we compare \ref{eq:FourierPerceptron} and \ref{eq:FourierSeriesVector}, we find similarities. We see that $ (a_\textbf{B},b_\textbf{B}) $ is equivalent to $ \textbf{W} $, $ \textbf{b} $ is zero and $\textbf{B}=\mathbb{N}_0\times \mathbb{Z}^{ d-1}, $ is the concatenation of all possible permutations of $ \textbf{n} $.  For practicality we limit \textbf{B} to
\begin{align}\label{def:B}
	\begin{split}
		\textbf{B}=\{0,\dots,N\} \times \{-N,\dots,N\}^{d-1}
		\setminus \mathbf{H},
	\end{split}
\end{align}
where $N$ will be called the frequency of the mapping, $\mathbf{H}=\{\textbf{n}\in \mathbb{N}_0\times \mathbb{Z}^{ d-1}|\, \exists j \in \{2,\dots,d\}:\, n_1=\dots=n_{j-1}=0 \wedge n_j<0\}$, then the perceptron represents a Fourier series. Hence, we calculate the dimension $m$ of all possible permutations (see supplementary material)
\begin{align}\label{mformel}
	m = (N+1)(2N+1)^{d-1}- \sum_{l=0}^{d-2}N(2N+1)^l .
\end{align} 
In practice, we can find the Fourier series coefficients by sampling the function uniformly with a frequency higher than the Nyquist frequency and apply a Fast Fourier Transform (FFT) on the sampled signal. The resulting FFT coefficients are the Fourier series coefficients multiplied by the number of the sampled points. And in theory, if we initialize the weights with the Fourier series coefficients, our network should give the training target at iteration 0. 
\subsection{SIRENs and Fourier mapping comparison}
\label{sec:SIRENs}
In this section we want show that a Fourier mapped perceptron is structurally like a SIREN with one hidden layer. If we evaluate $  \textbf{W}\cdot  \gamma(\textbf{x})$ in equation \eqref{eq:FourierPerceptron}, using \eqref{eq:fouriertransform} and combine the Sine and Cosine terms, we get:
\begin{align}\label{sirenform}
	\textbf{y}(\textbf{x}, \textbf{W}) =   \textbf{W}\cdot sin(2\pi \textbf{C}\cdot \textbf{x} + \phi) + \textbf{b},
\end{align}
where $\phi:=(\pi/2,\dots,\pi/2,0,\dots,0)^T \in \mathbb{R}^{2m}$ and $\textbf{C}:= (\textbf{B},\textbf{B})^T$ (see supplementary material).
Here we see that $ \textbf{C} $ is acting as the weight matrix applied to the input, $ \phi $ is like the first bias vector and $ sin(\cdot) $ is the activation function. Hence, the initial Fourier mapping can be represented by an extra initial SIREN layer,  with the difference that $ \textbf{B} $ and $ \phi $ are trainable in the SIREN case. This finding closes the bridge between Fourier frequency mappings and sinusoidal activation functions which have recently attracted a lot of attention.

\subsection{Progressive training}
\label{sec:progressive}
Chen-Hsuan et al. \cite{lin2021barf} introduced a training strategy for coarse-to-fine registration for NeRFs which they called BARF. Their idea is to mask out the positional encoding's high-frequency activations at the start of training and gradually allow them during training. Their work showed how to use this strategy on positional encodings only to improve camera registration. In our work, we will show how to run this strategy on an arbitrary Fourier mapping and show that it improves generalization of the interpolation task. We weigh the frequencies of $\gamma$ as follows:
\begin{align}
	\gamma^\alpha(\textbf{x}) := \left(\begin{array}{ccc}
		w_\textbf{B}^\alpha \\
		w_\textbf{B}^\alpha \\
	\end{array}\right)\circ \gamma(\textbf{x})
\end{align}
where $w_\textbf{B}^\alpha $ is the element wise application of the function $ w_\alpha(z)$ on the vector of Norms of $\textbf{B}$ on the input dimension:
\begin{align}
w_\textbf{B}^\alpha:=w_\alpha \left(\begin{array}{ccc}
||B_1||_2 \\
\vdots\\
||B_m||_2\\
\end{array}\right).
\end{align}
where $w_\alpha(z)$ is defined as:
\begin{align}
w_\alpha(z)= \begin{cases}
0  &\text{ if } \alpha - z  <0\\
\frac{1 - \cos((\alpha - z)\pi)}{2} &\text{ if } 0\le \alpha - z \le 1 \\
1 & \text{ if } \alpha - z > 1\end{cases}
\end{align}

Here, $ \alpha \in [0,\max((||B_i||_{d_\text{in}})_{i \in \{1,\dots,m\}})]$ is a parameter which is linearly increased during training. This strategy forces the network to train the low frequencies at the start of training, ensuring that the network will produce smooth outputs. Later, when high-frequency activations are allowed, the low-frequency components are trained, and the network can focus on the left details. This strategy should reduce the effect of overfitting, which was introduced by Tancik et al. \cite{tancik2020fourier} when using mappings with large standard deviations.

%

\subsection{Pruning}
\label{sec:pruning}
The standard way of using equation \eqref{def:B} is by defining a value $N$ and taking the whole set $\textbf{B}_N$. High-dimensional tasks lead to high memory consumption, and it is not clear whether this subset of $\mathbb{Z}^d$ brings best performance. We, therefore, propose a way to select a more appropriate subset through data pruning. A pruning $pr(N,M)$ is done as follows: Assume we have $N,M \in \mathbb{N}$ with $M>>N$ and $|\textbf{B}_N| = n$, $|\textbf{B}_M| = m$. We train a perceptron with an integer mapping given by $\textbf{B}_M$. After training we define $\textbf{D}$ such that $\textbf{D}$ contains only those elements of $\textbf{B}_M$ where the respective weights are greater than a margin, that is chosen to yield $|\textbf{D}| =n$. While $\textbf{B}_N$ and $\textbf{D}$ now have the same size, we believe that $\textbf{D}$ will yield better performance because it contains the most relevant frequencies of the signal we want to reconstruct. 
\begin{figure*}[h]
	\begin{subfigure}{0.19\textwidth}
		\centering
		\includegraphics[width=\textwidth]{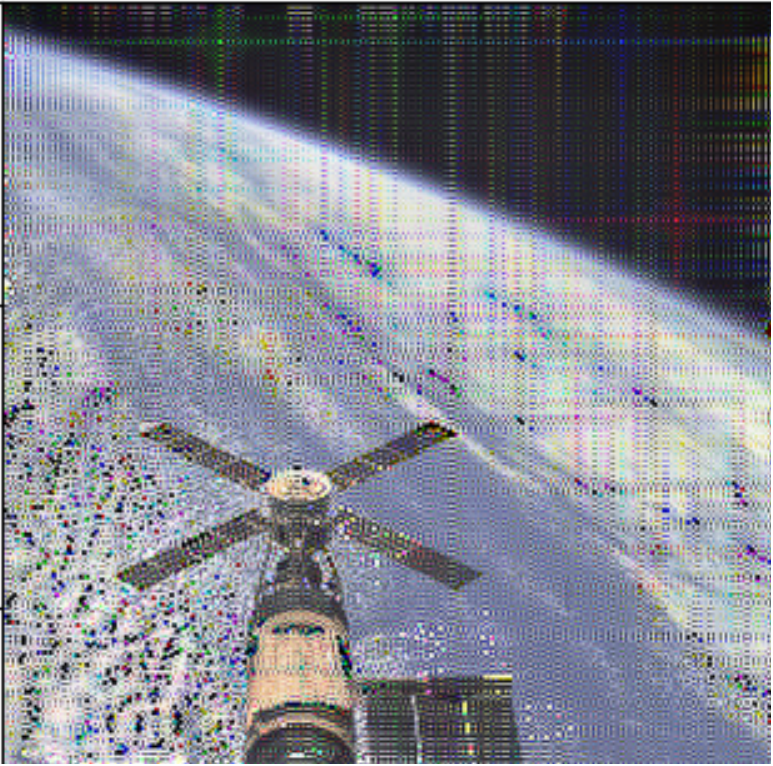}
		\caption{PT=F, WI=F}
		\label{fig:ver_FF}
	\end{subfigure}
	\begin{subfigure}{0.19\textwidth}
		\centering
		\includegraphics[width=\textwidth]{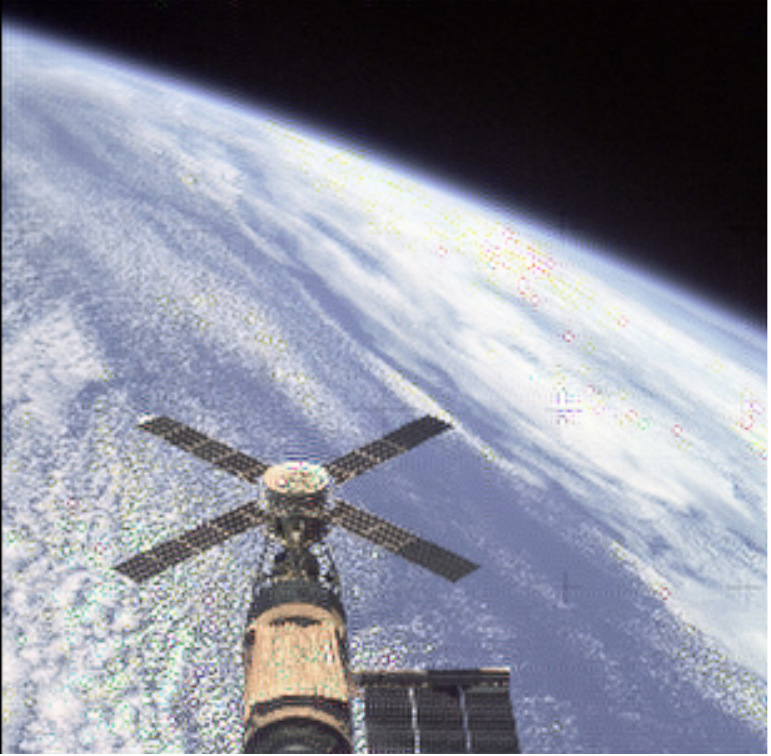}
		\caption{PT=F, WI=T}
		\label{fig:ver_FT}
	\end{subfigure}
	\begin{subfigure}{0.19\textwidth}
		\centering
		\includegraphics[width=\textwidth]{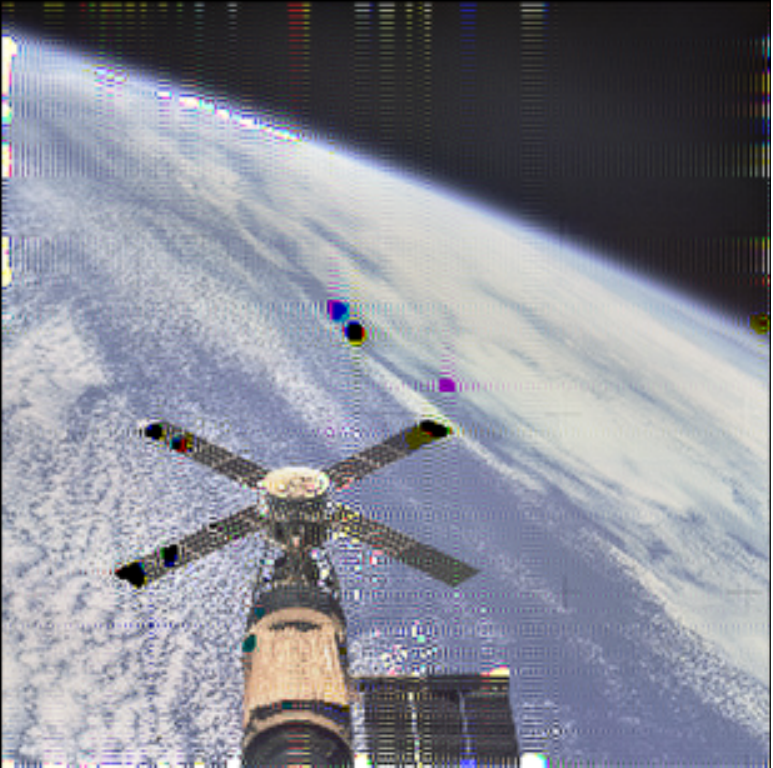}
		\caption{PT=T, WI=F}
		\label{fig:ver_TF}
	\end{subfigure}
	\begin{subfigure}{0.19\textwidth}
		\centering
		\includegraphics[width=\textwidth]{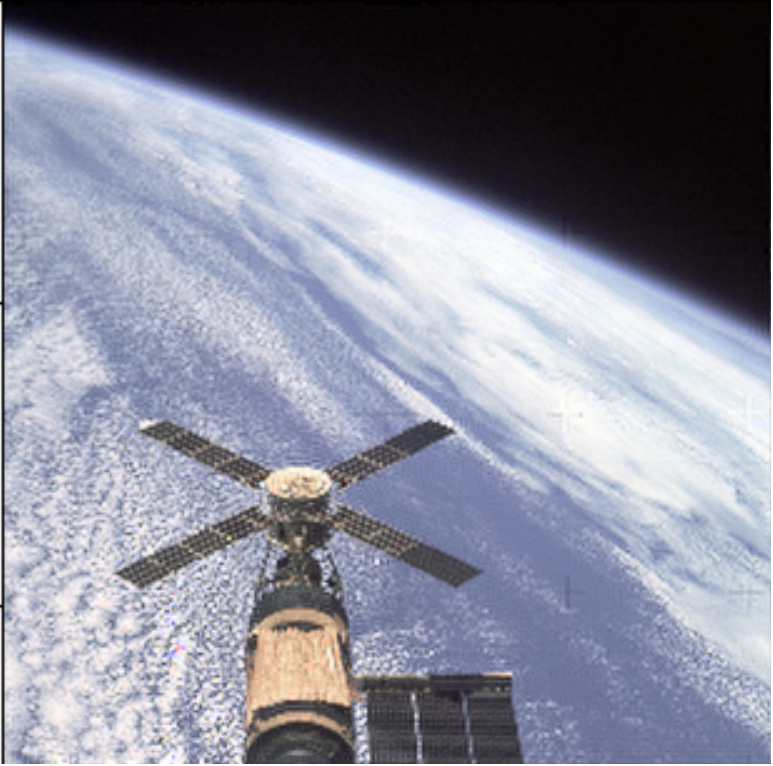}
		\caption{PT=T, WI=T}
		\label{fig:ver_TT}
	\end{subfigure}
	\begin{subfigure}{0.19\textwidth}
		\centering
		\includegraphics[width=\textwidth]{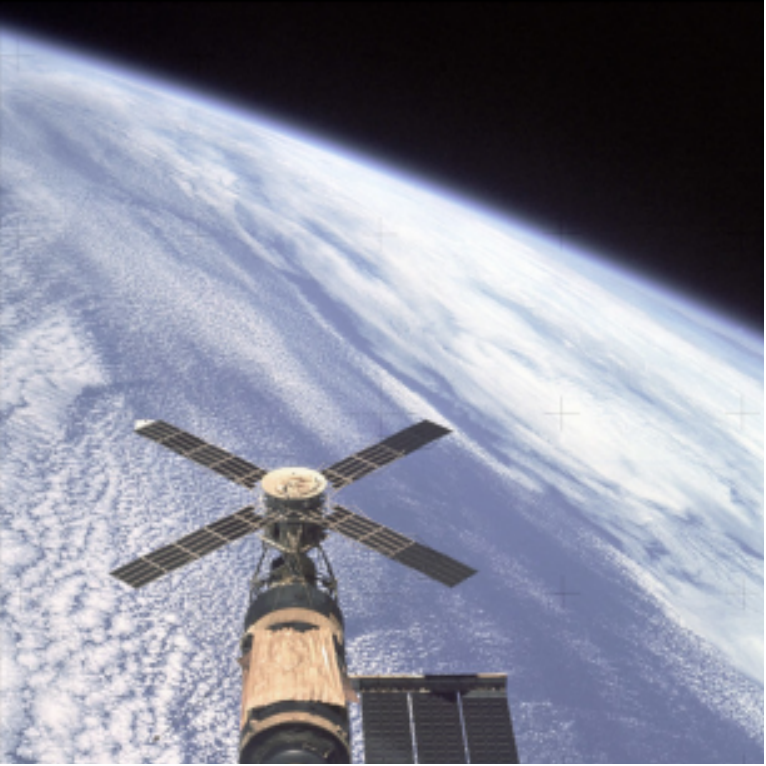}
		\caption{GT}
		\label{fig:ver_GT}
	\end{subfigure}
	\caption{A visualization of the outputs of Fourier mapped perceptrons of $N = 128$. PT stands for progressive training and WI stands for weight initialization. T/F stands for True/False, respectively. }
	\label{fig:verifications}
\end{figure*}
\begin{figure}
	\begin{center}
		\includegraphics[trim={1cm 0cm 4cm 1cm},clip,width=0.235\textwidth]{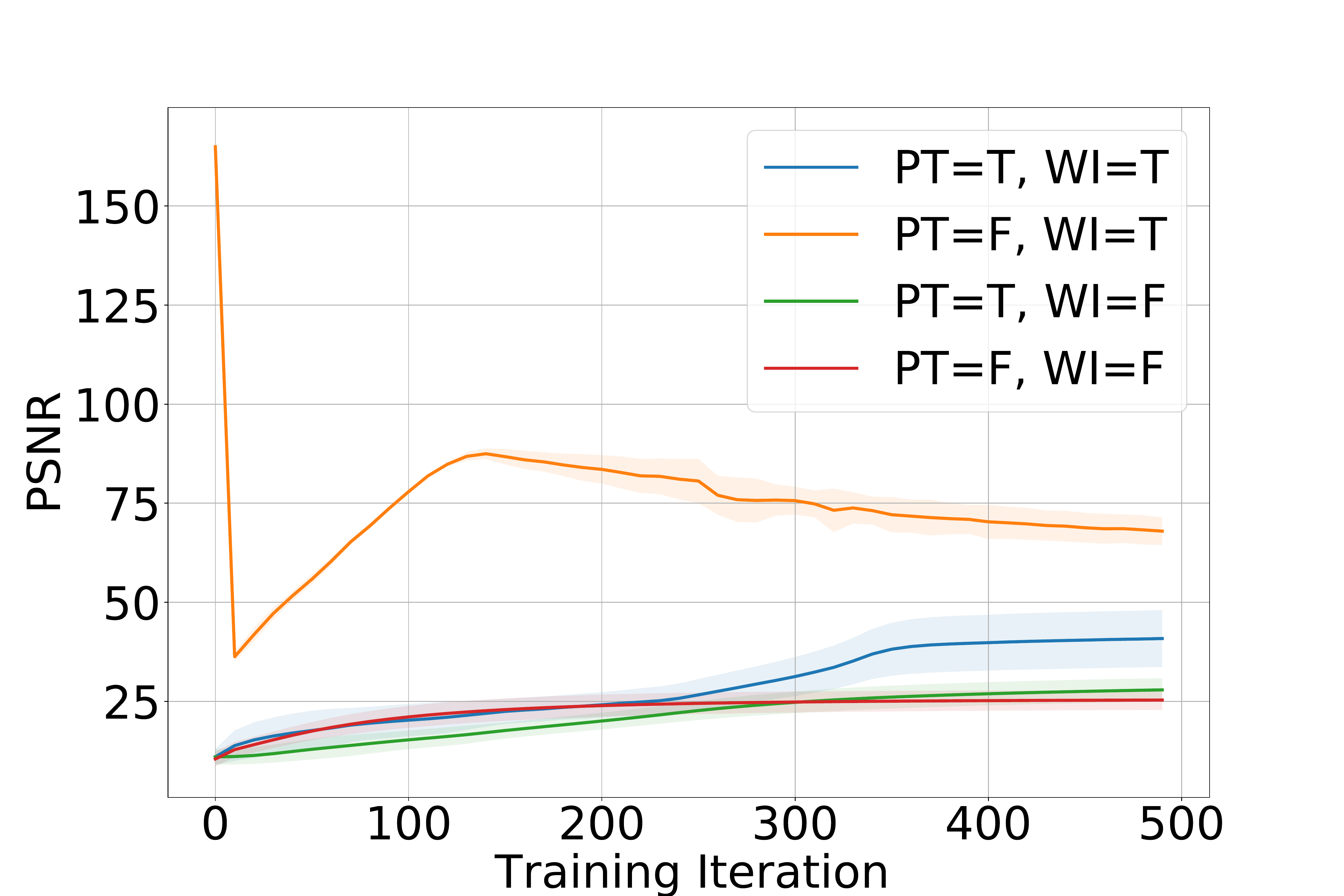}
		\includegraphics[trim={2cm 0cm 4cm 1cm},clip,width=0.235\textwidth]{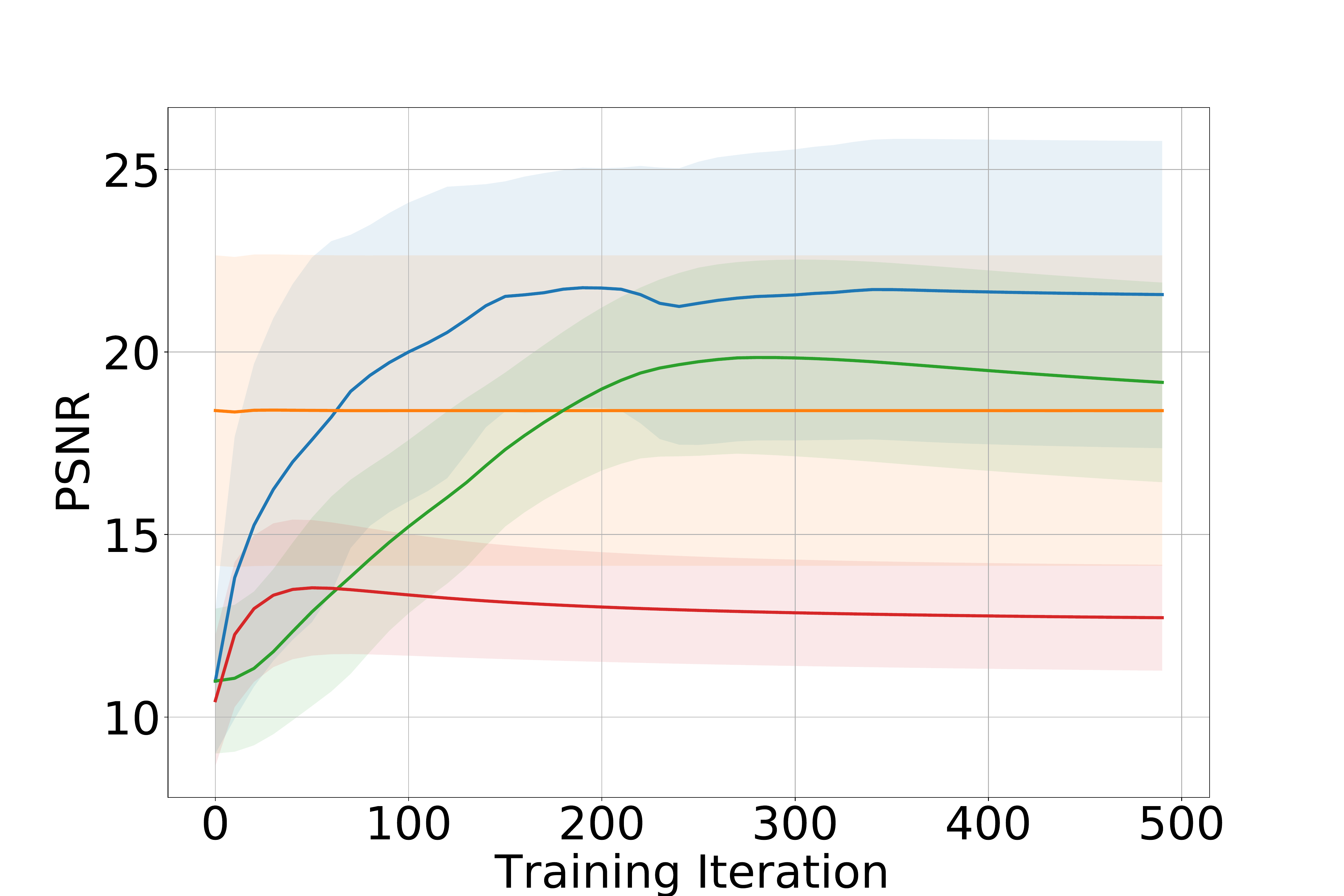}
	\end{center}\caption{The training progress of Fourier mapped perceptrons with $ N = 128$. The left and the right figures report the train and test PSNR, respectively.  Weight initialization without PT yields a PSNR of 160 which one can consider as the ground truth proving that the perceptron is a Fourier series. Note: The y-axis limits are different in both plots.}
	\label{fig:weight_int}
\end{figure}
\begin{figure*}
	\centering
	\begin{subfigure}[b]{0.4\textwidth}
		\centering
		\includegraphics[trim={0.5cm 0cm 2cm 1.5cm},clip,width=\textwidth]{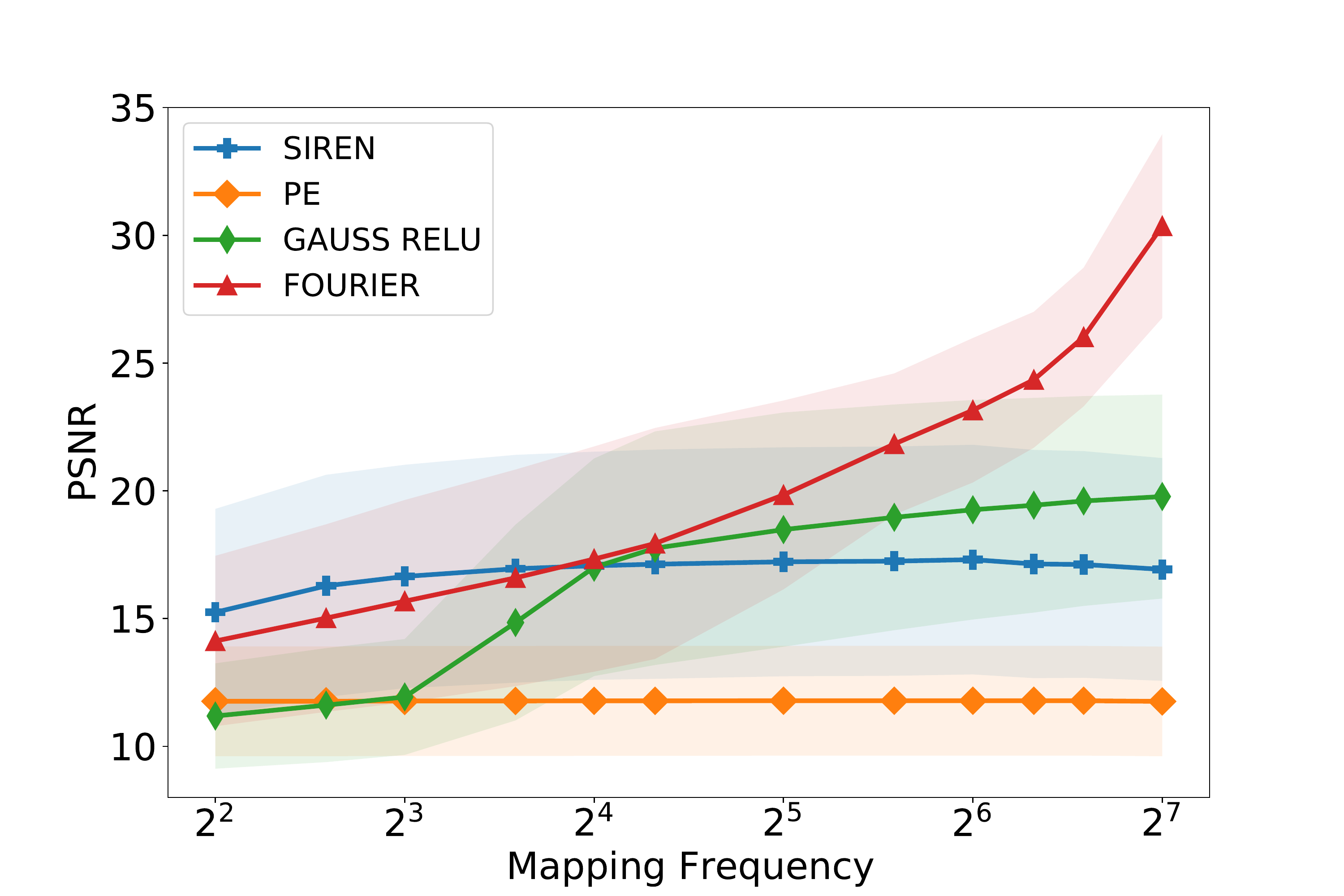}
		\label{fig:perceptron_train}
	\end{subfigure}
	\begin{subfigure}[b]{0.4\textwidth}
		\centering
		\includegraphics[trim={0.5cm 0cm 2cm 1.5cm},clip,width=\textwidth]{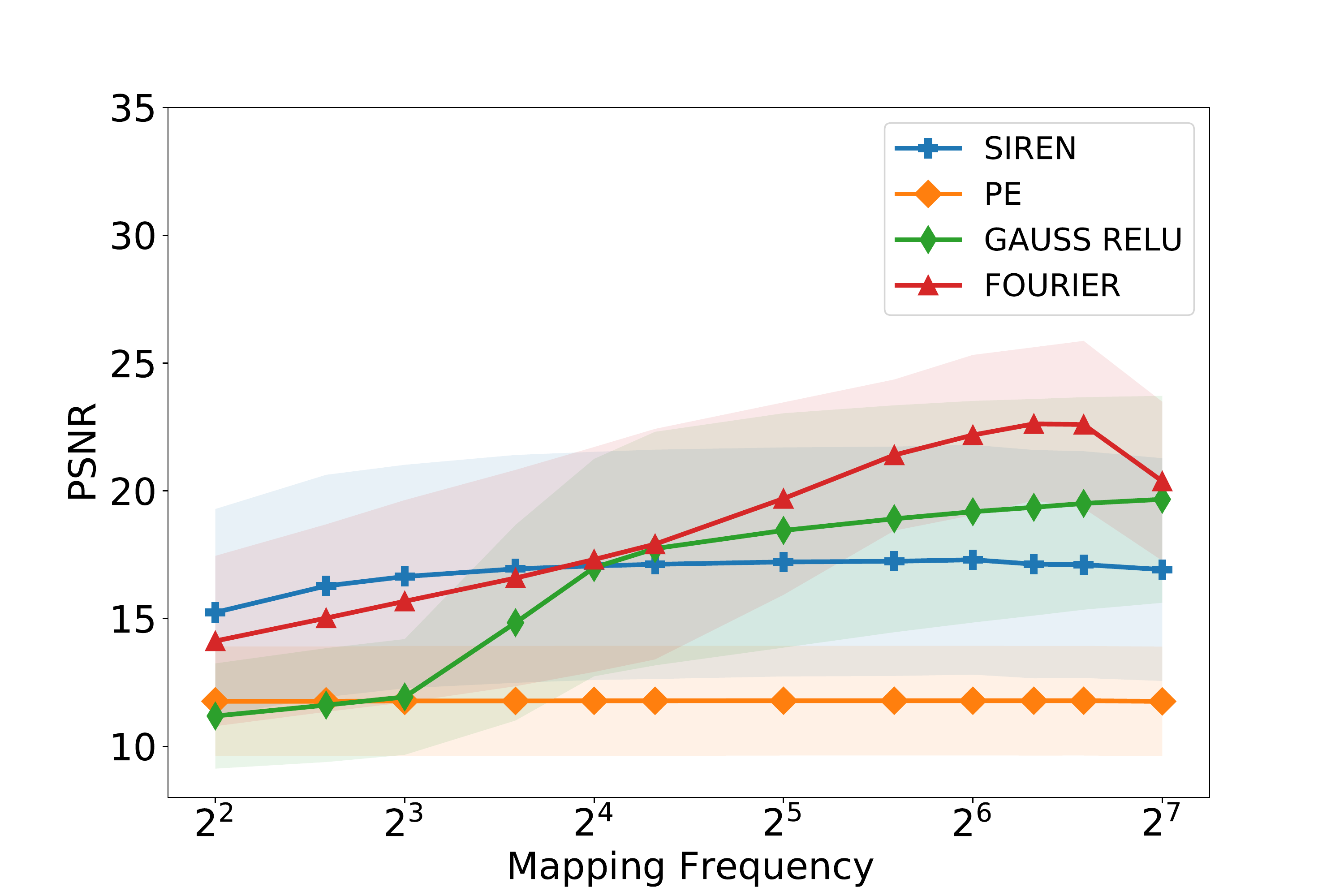}		
		\label{fig:perceptron_test}
	\end{subfigure}
	\caption{Perceptron experiments with different values for the mapping frequency $N$. We report the train PSNR on the left and the test PSNR on the right. For high values of $N$ our integer mapping outperforms all competing mappings.}
	\label{fig:perceptron}
\end{figure*}
\begin{figure*}	
	\begin{subfigure}{\textwidth}
		\centering
		\includegraphics[trim={6cm 10.2cm 5cm 1cm},clip,width=\textwidth]{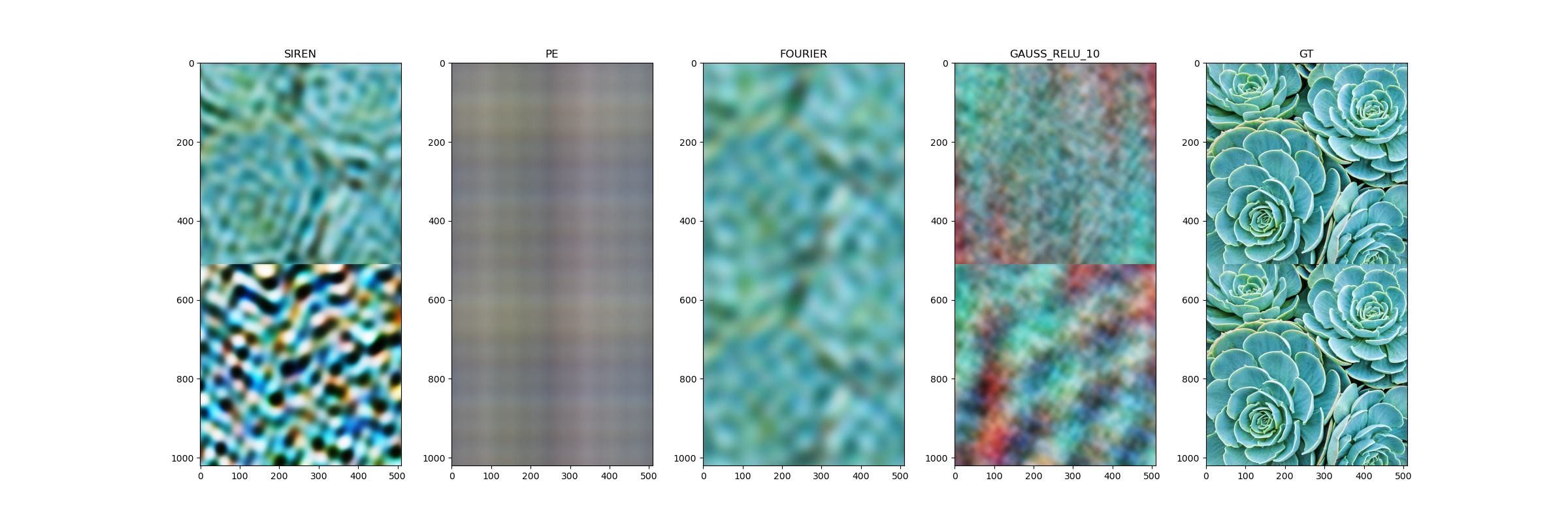}
		\caption{Network predictions using $ N $ = 8}
		\label{fig:vis8}
	\end{subfigure}
	\begin{subfigure}{\textwidth}
		\centering
		\includegraphics[trim={6cm 10.2cm 5cm 1.5cm},clip,width=\textwidth]{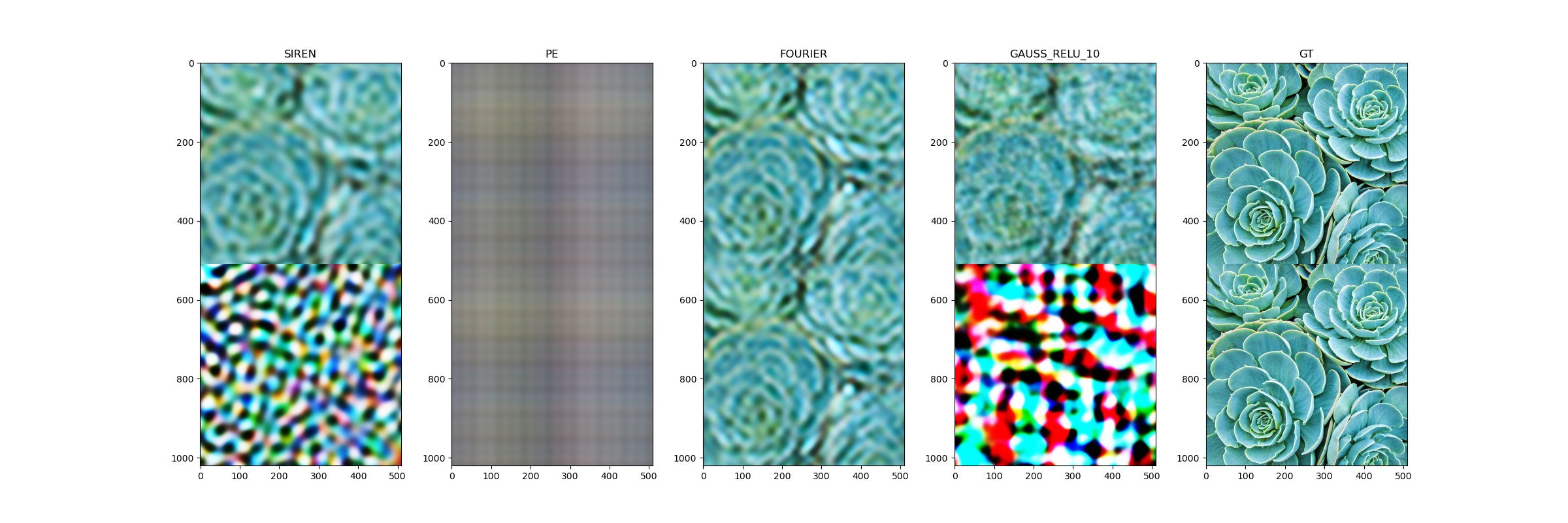}
		\caption{Network predictions using $ N $ = 16}
		\label{fig:vis16}
	\end{subfigure}
	\begin{subfigure}{\textwidth}
		\centering
		\includegraphics[trim={6cm 10.2cm 5cm 1.5cm},clip,width=\textwidth]{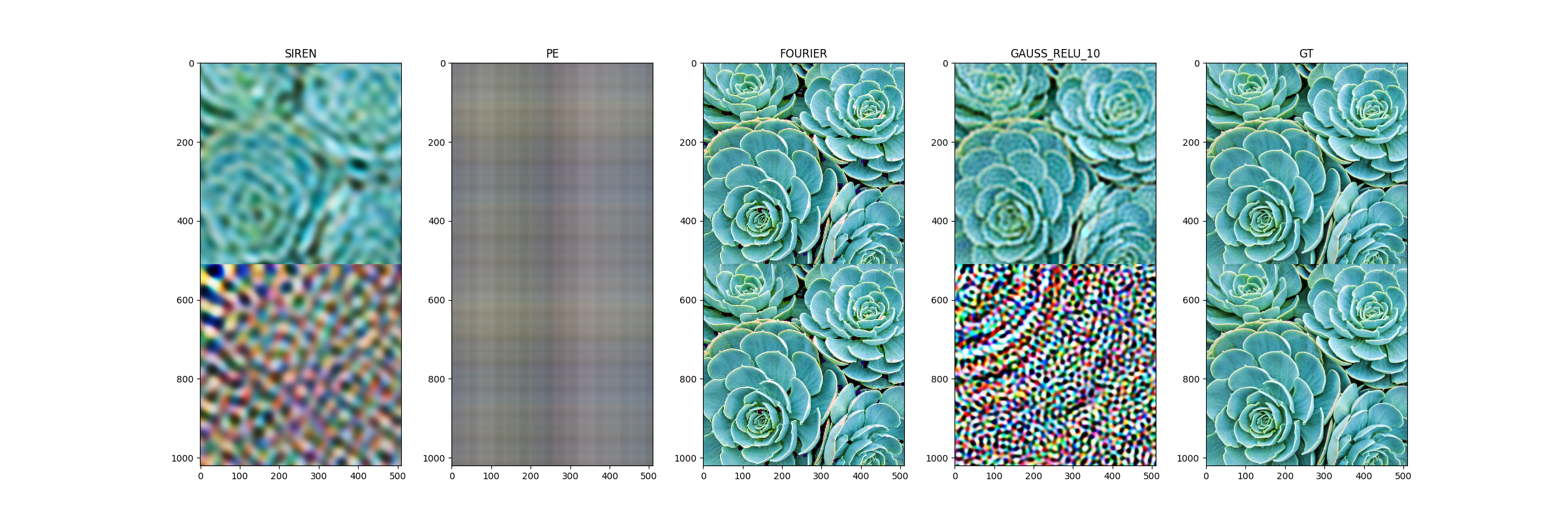}
		\caption{Network predictions using $ N $ = 128}
		\label{fig:vis96}
	\end{subfigure}
	
	\caption{The visualization of the Fourier mapped perceptrons and the one layer SIREN with different values of $N$.}
	\label{fig:visulazation}
\end{figure*}

\subsection{Integer lattice mapping applied to MLPs}
\label{sec:int effectiveness}

Although we showed in section \ref{sec:integer lattice mapping} that we can represent any bounded input function with only one Fourier mapped perceptron, in practice, these networks can become very wide to give high performance. As a result, the number of calculations will increase. To compromise between performance and speed, one can add depth and reduce the width of the network.

First, it is natural that using MLPs rather than perceptrons increases the performance. However, it remains unclear why our proposed integer mapping should perform better than competing mappings for multilayer networks. 

One could argue that if a mapping gives the perceptron a high representation power, it will also provide a high representation power to the MLP and vice versa. First, however, we should verify this claim with experiments.
In addition, we remind the reader that a periodic function has integer frequencies. And because our assumption that the signal we want to reconstruct is periodic, it will have only integer frequencies. Also, the activation functions we are using only introduce integer frequencies when applied to a periodic function, as shown for the 1D case in the supplementary material. With this, we reduce the search space for frequencies from $\mathbb{R}$ to $\mathbb{Z}$, which could make the optimization easier as the search space is more compact and approachable.

\section{Experiments}


\subsection{Weight initialization and progressive training}
\label{sec:WIandPT}
In this section, we want to prove our mathematical claims by experiments. First, we will show that the derivation of the integer mapping indeed represents the Fourier series. Secondly, we want to check whether progressive training helps with generalization.

We conducted our experiments on the image regression task. This task aims to make a neural network memorize an image by predicting the color at each pixel location. We use ten images with a resolution of $512 \times 512 $, which can be found in the supplementary material, and report the mean peak signal-to-noise ratio (PSNR). We divide the image into train and test sets, where we use every second pixel for training and take the complete image for testing. We utilize 3 Fourier-mapped perceptrons with $ N = 128$ (Nyquist frequency), one for each image channel. We normalize the input ($\textbf{x}$) to have an interval between [0,1] in both width ($x$) and height ($y$) dimensions. 

In this experiment, we made an ablation: With and without weight initialization using the normalized FFT coefficients of the image's training pixels, with and without the progressive training scheme explained in section \ref{sec:progressive}. For progressive training, $ \alpha$ was linearly increased from 0 to its maximum value at 75\% of training iterations. In training, we only make an update step after we accumulate the gradients of the whole image. We did not study learning schedules in this work, and the reader is encouraged to try different schedules. Figure \ref{fig:verifications} shows a visualization of one of the images, and figure \ref{fig:weight_int} shows the training progress, where the solid line is the mean PSNR and the shaded area shows the standard deviation.

As can be deducted from figure \ref{fig:weight_int}, one can see that the train PSNR starts at an optimum at the start of training when we use weight initialization (WI), and we don't use progressive training (PT). This fact underlines our claim that \textit{a perceptron with an integer lattice mapping is indeed a Fourier series}. Note that in case both WI and PT are used, the train PSNR is not optimal at the start because the PT masks out high-frequency activations.

We can also see from figure \ref{fig:weight_int} that whenever we use progressive training, it always shows a higher test PSNR, which certifies that \textit{progressive training helps with generalization}. Lastly, when we did not employ both PT and WI, the perceptron overfits to the training pixels, and this can be seen quantitatively with a very low test PSNR (red line in figure \ref{fig:weight_int}) and qualitatively with grid-like artifacts (in figure \ref{fig:ver_FF})).

\begin{table*}[]
\begin{tabular}{|c|c|c|c|c|c|c|c|c|c|c|c|c|c|}
	\hline
	\multirow{2}{*}{Activ.} & \multirow{2}{*}{Map.} & \multicolumn{4}{c|}{$N$=8 $m$=113}                                  & \multicolumn{4}{c|}{$N$=16 $m$=481}                                 & \multicolumn{4}{c|}{$N$=32 $m$=1985}                                \\ \cline{3-14} 
	&                       & d=0            & d=2            & d=4            & d=6            & d=0            & d=2            & d=4            & d=6            & d=0            & d=2            & d=4            & d=6            \\ \hline
	\multirow{3}{*}{Sine}   & No                    & \textbf{16.65} & 22.15          & \textbf{23.26} & \textbf{24.07} & 17.07          & 22.09          & 23.84          & 19.76          & 17.22          & 14.90          & 14.67          & 13.63          \\ \cline{2-14} 
	& Int.               & 15.68          & \textbf{22.31} & 22.41          & 20.94          & \textbf{17.33} & 27.66          & 27.06          & \textbf{27.33} & \textbf{19.84} & 33.78          & 26.98          & 23.60          \\ \cline{2-14} 
	& Pr.                 & 15.28          & 21.03          & 22.40          & 23.00          & 16.76          & \textbf{28.17} & \textbf{27.68} & 24.66          & 18.48          & 37.34          & 30.41          & 19.74          \\ \hline
	\multirow{4}{*}{Relu}   & P.E.                  & 11.78          & 16.61          & 17.37          & 17.77          & 11.78          & 16.87          & 17.79          & 17.95          & 11.78          & 17.05          & 18.15          & 18.15          \\ \cline{2-14} 
	& Gs. $\sigma_{10}$  &   11.93             & 21.90          & 21.68          & 21.69         &     17.01           & 24.53          & 24.26          & 25.13          &   18.48             & 26.10          & 26.30 & 27.48 \\ \cline{2-14} 
	& Gs. $\sigma_{pr}$      & 14.06   & 20.23                   & 20.78                  & 20.88                    & 12.69    & 26.02                    & 26.40                  & 26.72                    & 13.01    & 37.69                    & \textbf{37.90}                    & \textbf{37.74}                    \\ \cline{2-14} 
	& Int.               & 15.68          & 20.51          & 20.65          & 20.62          & \textbf{17.33} & 24.42          & 24.09          & 24.49          & \textbf{19.84} & 31.57          & 32.14          & 32.79          \\ \cline{2-14} 
	& Pr.                 & 15.28          & 20.35          & 20.92          & 20.96          & 16.76          & 25.87          & 26.23          & 26.33          & 18.48          & \textbf{37.70} & 36.81          & 37.48          \\ \hline
\end{tabular}
\caption{The mean train PSNR results of network type comparison experiment with varying network depth (d), number of frequencies ($N$). We use the following abbreviations: Activ. = Activation function, Map. = Mapping type, Int. = Integer, Pr.= Pruned Integer, P.E. = Positional Encoding, Gs. = Gaussian. Here, $m$ is the mapping size and $\sigma$ is the standard deviation.}
\label{tab:mlp_train}
\end{table*}

\begin{table*}[]
\begin{tabular}{|c|c|r|r|r|r|r|r|r|r|r|r|r|r|}
	\hline
	\multirow{2}{*}{Activ.} & \multirow{2}{*}{Map.} & \multicolumn{4}{c|}{$N$=8 $m$=113}                                                                          & \multicolumn{4}{c|}{$N$=16 $m$=481}                                                                         & \multicolumn{4}{c|}{$N$=32 $m$=1985}                                                                        \\ \cline{3-14} 
	&                       & \multicolumn{1}{c|}{d=0} & \multicolumn{1}{c|}{d=2} & \multicolumn{1}{c|}{d=4} & \multicolumn{1}{c|}{d=6} & \multicolumn{1}{c|}{d=0} & \multicolumn{1}{c|}{d=2} & \multicolumn{1}{c|}{d=4} & \multicolumn{1}{c|}{d=6} & \multicolumn{1}{c|}{d=0} & \multicolumn{1}{c|}{d=2} & \multicolumn{1}{c|}{d=4} & \multicolumn{1}{c|}{d=6} \\ \hline
	\multirow{3}{*}{Sine}   & No                    & \textbf{16.65}           & 21.63                    & \textbf{21.85}           & \textbf{21.99}           & 17.06                    & 21.28                    & 22.03                    & 18.50                    & 17.22                    & 13.57                    & 13.29                    & 12.37                    \\ \cline{2-14} 
	& Int.               & 15.68                    & \textbf{21.75}           & 21.53                    & 20.06                    & \textbf{17.31}           & \textbf{23.48}           & \textbf{22.67}           & 22.28                    & \textbf{19.70}           & 16.85                    & 17.89                    & 16.36                    \\ \cline{2-14} 
	& Pr.                 & 15.28                    & 20.49                    & 21.22                    & 21.45                    & 16.75                    & 22.00                    & 21.39                    & 22.17                    & 18.39                    & 20.49                    & 15.15                    & 13.13                    \\ \hline
	\multirow{4}{*}{Relu}   & P.E.                  & 11.78                    & 16.60                    & 17.33                    & 17.70                    & 11.78                    & 16.85                    & 17.73                    & 17.87                    & 11.79                    & 17.02                    & 18.06                    & 18.02                    \\ \cline{2-14} 
	& Gs. $\sigma_{10}$                 & 11.93   & 20.67                    & 21.06                    & 20.90                    & 17.00    & 22.96                    & 22.78                    & 23.04                    & 18.45    & 23.66                   & 23.61                    & \textbf{23.73}                    \\ \cline{2-14} 
	& Gs. $\sigma_{pr}$                 & 14.06   & 19.89                    & 20.22                    & 20.21                    & 12.69    & 22.46                    & 22.48                    & 22.16                    & 12.99    & 23.12                    & 23.48                    & 23.33                    \\ \cline{2-14} 
	& Int.               & 15.68                    & 20.27                    & 20.35                    & 20.23                    & \textbf{17.31}           & 22.93                    & 22.65                    & \textbf{22.50}           & \textbf{19.70}           & \textbf{24.36}           & \textbf{24.02}           & \textbf{23.73}           \\ \cline{2-14} 
	& Pr.                 & 15.28                    & 19.98                    & 20.33                    & 20.21                    & 16.75                    & 22.31                    & 22.26                    & 22.09                    & 18.39                    & 23.24                    & 23.18                    & 23.30                    \\ \hline
\end{tabular}
\caption{The mean test PSNR results of network type comparison experiment. For abbreviations see table \ref{tab:mlp_train}. }
\label{tab:mlp_test}
\end{table*}

\subsection{Perceptron experiments}
\label{sec:perceptron experiments}
In this experiment, we want to compare the representation power of the different mappings in the single perceptron case. We conducted our experiments in the same setting as in section \ref{sec:WIandPT}, where we used progressive training and did not use weight initialization.

In the integer mapping, we increased $ N $'s value from 4 until half the training image dimension (Nyquist frequency) and calculated all possible permutations $ \textbf{B}_N$, as discussed in section \ref{sec:integer lattice mapping}. For the Gaussian mapping, we sample $ m=|\textbf{B}_N|$ parameters from a Gaussian distribution with a standard deviation of 10 (which was the best value for this task in our experiments). Also, we test a one-layer SIREN with one hidden layer having the same size $m$. Finally, we adopt the positional encoding (PE) scheme from \cite{mildenhall2020nerf} and limit its values to $N$. Figure \ref{fig:perceptron} shows our experiments' results on the train and test pixels, respectively. Figure \ref{fig:visulazation} shows the networks' outputs trained on one of the images.

At low N values (figure \ref{fig:vis8}), we see that the Gaussian mapped perceptrons do not work because the number of sampled frequencies is low, so there is a low chance that samples will be near the image's critical frequencies. On the other hand, the integer mapped perceptrons give a blurry image because they can only learn low frequencies. The SIREN performs relatively well in this case, and we think this is because SIRENs naturally inherit a learnable Fourier mapping that is not restricted to the initial sampling, as described in section \ref{sec:SIRENs}. PE can only produce horizontal and vertical lines because it has diagonal frequencies (only one non-zero frequency is allowed), and this effect is persistent at any value of $N$.

As $N$ increases, SIREN, Gauss, and integer mapping performance increase giving similar performance around $N=16$ (figure \ref{fig:vis16}). For high values of $N$, we see that in figure \ref{fig:vis96}, the integer lattice mapping of the Fourier coefficients outperforms the competing mappings, clearly displaying more details in the reconstruction. On the other hand, the PSNR of the SIREN and the Gaussian mapped perceptrons saturates. We think this is because both mappings rely on sampling the frequencies. Although we can get many of the critical frequencies of the image with sampling, it is improbable to get all of them simultaneously. Even the trainability of the SIREN mapping did not help in this case.
\begin{figure*}
	\begin{subfigure}{.19\linewidth}
		\centering
		\includegraphics[width=0.95\linewidth]{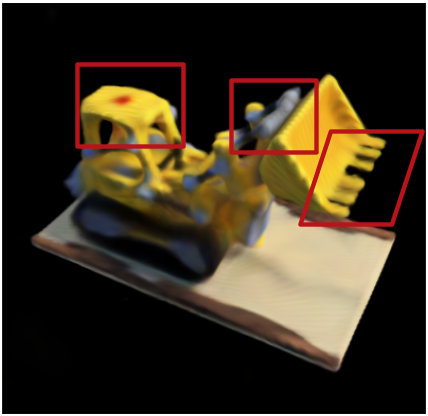}
		\caption{ReLu without pruning.}
		\label{fig:real}
	\end{subfigure}
	\begin{subfigure}{.19\linewidth}
		\centering
		\includegraphics[width=0.95\linewidth]{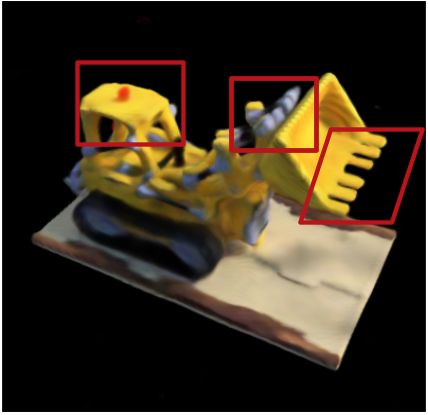}
		\caption{ReLu with pruning.}
		\label{fig:pose_all}
	\end{subfigure}
	\begin{subfigure}{.19\linewidth}
		\centering
		\includegraphics[width=0.95\linewidth]{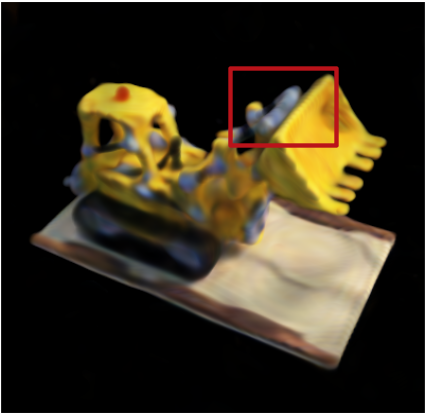}
		\caption{Sine without pruning.}
		\label{fig:det}
	\end{subfigure}
	\begin{subfigure}{.19\linewidth}
		\centering
		\includegraphics[width=0.95\linewidth]{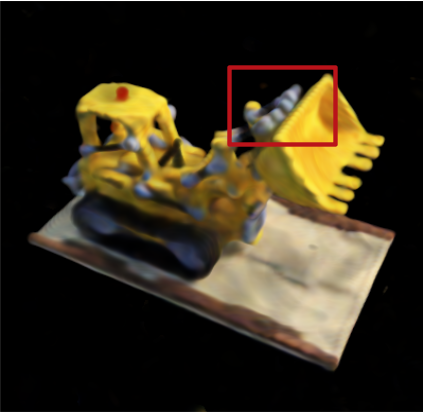}
		\caption{Sine with pruning.}
		\label{fig:pose_5}
	\end{subfigure}
	\centering
	\begin{subfigure}{.19\linewidth}
		\centering
		\includegraphics[width=0.95\linewidth]{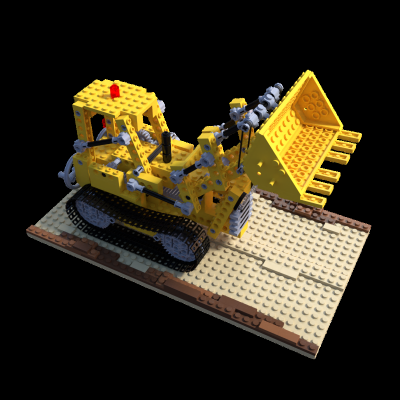}
		\caption{GT}
		\label{fig:pose_6}
	\end{subfigure}
	\caption{View synthesis results using a simplified Nerf. A small MLP with a depth of 4, width of 64 and integer mapping with a frequency of 4 is used. The pruning is done with pr(4,8). The pruning technique shows qualitative improvements.}
	\label{fig:Nerf}\vspace{-0.2cm}\vspace{-0.2cm}
\end{figure*}

\subsection{MLP experiments}
\label{sec:mlp}
Our theory for integer mapping assumes an underlying function that is periodic. However, it is not clear that we will end up with a periodic function if we go the other way, using an integer mapping. In this experiment, we want to check if applying an integer mapping forces periodicity. Secondly, we want to validate our claim (in section \ref{sec:int effectiveness}) that if a mapping gives the perceptron a high representation power, it will also give a high representation power to the MLP and vice versa. We compared ReLU networks with integer, Gaussian, PE, and pruned integer mapping (section \ref{sec:pruning}). We also compared SIRENS with no mapping (extra layer), integer, pruned mapping. We made a grid search of the parameters N=[8, 16, 32], depth=[0, 2, 4, 6] (depth=0 represents a perceptron), and fixed the width to 32. For the pruned mapping, we used a $pr(N, 128)$. And for the Gaussian mapping, we had two settings. The first one had a standard deviation of 10 ($\sigma_{10}$), which had the best performance in the perceptron experiments. In the second one, we set the standard deviation the same as the pruned integer mapping's standard deviation ($\sigma_{pr}$) to check its effect. Tables \ref{tab:mlp_train} and \ref{tab:mlp_test} show the mean train and test PSNRs respectively.

Figure \ref{fig:main} shows a visualization of the network's outputs at N = 16, depth = 4 and width = 32 for the first period and next period in the height and width directions ($ f([x + 1, y + 1 ]) $). And we see that \textit{the integer mapping forces the network's underlying function to be periodic} unlike the SIREN and ReLU network with Gauss mapping, which proves our first hypothesis.  

From the table \ref{tab:mlp_train} we see that if a mapping at $d=0$ gives the highest PSNR, this does not mean that it will give the highest PSNR for $d>0$ and vice versa. One clear example at $N=32$ is the Gauss $\sigma_{pr}$, where it has a PSNR of 13.01\,dB  at $d=0$, which is lower than integer mapping (19.84\,dB), but has the highest PSNR at $d=[4,6]$. This result disproves our initial assumption that if a mapping gives the perceptron a high representation power, it will also give a high representation power to the MLP. We see also that the pruned integer mapping has comparable results with the Gauss $\sigma_{pr}$, and this shows that the main contributor to the performance is the mappings' standard deviation.

From the tables, we can also observe some trends. First, networks with sine activations and large mappings collapse during perform worse than Relu networks. Second, the integer mapping usually gives the best test PSNR, demonstrating its effectiveness in the MLP case. Third, the pruned integer mapping shows consistently better train PSNR than the normal integer mapping at $d>0$. We believe this is because pruned mapping has a higher standard deviation. Finally, the PE is worse in every case because we cannot easily control the standard deviation, and it has very few parameters.

\begin{table}[]
\begin{tabular}{|c|c|c|c|c|c|c|}
	\hline
	\multirow{2}{*}{Act.} & \multirow{2}{*}{Map.} & \multicolumn{4}{c|}{$N$ = 4}                                      & $N$=8          \\ \cline{3-7} 
	&                       & d=0            & d=2            & d=4            & d=6            & d=0            \\ \hline
	\multirow{3}{*}{Sine} & No                    & \textbf{20.37} & 23.08          & \textbf{23.55} & 23.35          & OM             \\ \cline{2-7} 
	& Int.                  & 18.42          & 22.22          & 22.95          & 22.97          & \textbf{19.31} \\ \cline{2-7} 
	& Pr.                   & 19.15          & \textbf{23.12} & \textbf{23.58} & 23.36          & -              \\ \hline
	\multirow{4}{*}{Relu} & P.E.                  & 16.30          & 21.48          & 22.64          & 23.51          & 16.40          \\ \cline{2-7} 
	& Gs.                   & 18.93          & 22.81          & \textbf{23.64} & \textbf{23.82} & 19.29          \\ \cline{2-7} 
	& Int.                  & 18.42          & 21.81          & 22.68          & 23.28          & \textbf{19.31} \\ \cline{2-7} 
	& Pr.                   & 19.15          & 22.78          & 23.61          & \textbf{23.89} & \textbf{-}     \\ \hline
\end{tabular}
\caption{Validation PSNR scores of Nerf experiments using a mapping of frequency 4. OM stands for out of memory. For other abbreviations see table \ref{tab:mlp_train}.}
\label{table:Nerf}
\end{table}

\subsection{Novel view synthesis experiments}
This section wants to see if our findings in the image regression task transfer to the novel view synthesis (NVS) task. In NVS, we are given a set of 2D images of a scene, and we try to find its 3D representation. With this representation, one can render images from new viewpoints. In contrast to the 2D experiments, the inputs are $(x,y,z)$ coordinates that are mapped to a 4-dimensional output, the RGB-values, and a volume density. For this experiment, a simplified version of the official NeRF \cite{mildenhall2020nerf} is used, where the view dependency and hierarchical sampling are removed. Here, we experiment with the input mappings used in section \ref{sec:mlp}. Unless otherwise stated, we adopt the settings from the image regression task. We set the network width to be 64.  

As the mapping size increases exponentially, we do our experiments with lower frequencies than in the 2D case. Specifically, we used the integer mapping on four frequencies. The frequencies of our mapping were limited to the maximum network size which we could fit on NVIDIA GTX-2080Ti. The pruning is given by $pr(4,8)$. We conduct our experiments on the bulldozer scene, which is commonly used for Nerf experiments. For training, we used a batch size of 128, 50.000 epochs and a learning rate of $5 \times 10^{-4}$. 

As seen in Table \ref{table:Nerf}, in the perceptron case ($d=0$), SIREN provides the best performance, which aligns with our image regression results at low values of $N$. We observe that the pruned mapping increases the performance compared to normal mapping for both Relu and sinusoidal activation. This increase in performance is because pruned mapping has a higher standard deviation than normal mapping. Qualitative improvements of the pruning can be seen in Figure \ref{fig:Nerf}. Gauss gives comparable results to pruned integer mapping because they have the same standard deviation. These findings align with our conclusions from image regression experiments. However, due to memory limitations, we could not test a perceptron with frequencies higher than 8, which was superior in image regression.

\section{Conclusion}
In this work, we identified a relationship between the Fourier mapping and the general d-dimensional Fourier series, which led to the \textit{integer lattice mapping}. We also showed that this mapping forces periodicity of the neural network underlying function. From experiments, we showed that one perceptron with frequencies equal to the Nyquist rate of the signal is enough to reconstruct it. Furthermore, we showed that the progressive training strategy improves the generalization of the interpolation task. Lastly, we confirmed the previous findings that the main contributor to the mapping performance is its size and the standard deviation of its elements.  

{\small
\bibliographystyle{ieee_fullname}
\bibliography{egbib}
}

\end{document}